\pdfoutput=1
\documentclass[11pt]{article}
\usepackage{acl}

\usepackage{times}
\usepackage{latexsym}
\usepackage{graphicx}
\usepackage{multirow}
\usepackage[inline]{enumitem}
\usepackage{amsmath}
\usepackage{amssymb,amsfonts}
\usepackage{float}
\usepackage{booktabs}
\usepackage{subcaption}
\usepackage[skip=2pt]{caption}
\usepackage[update,prepend]{epstopdf}
\usepackage[lined,algonl,ruled]{algorithm2e}
\usepackage{balance}
\usepackage{acronym}
\usepackage{color}
\usepackage{multirow}
\usepackage{multicol}
\usepackage{hyperref}
\usepackage{pifont}

\acrodef{SOTA}{State-of-The-Art}
\acrodef{CNN}{Convolutional Neural Network}
\acrodef{WER}{Word Error Rate}
\acrodef{CER}{Character Error Rate}
\acrodef{PER}{Phoneme Error Rate}
\acrodef{UER}{Unit Error Rate}
\acrodef{MAE}{Mean Absolute Error}
\acrodef{MSE}{Mean Squared Error}

\acrodef{ASR}{Automatic Speech Recognition}
\acrodef{TTS}{Text-to-Speech}
\acrodef{GMM}{Gaussian Mixture Model}
\acrodef{HMM}{Hidden Markov Model}
\acrodef{MOS}{Mean-Opinion-Score}
\acrodef{eMOC}{Emotion-Mean-Opinion-Classification}
\acrodef{NLP}{Natural Language Processing}
\acrodef{SSL}{Self-Supervised Learning}

\acrodef{S2u}{Speech-to-units}
\acrodef{u2S}{units-to-Speech}
\acrodef{uLM}{units-Language-Model}

\usepackage{bm}
\usepackage{amssymb,amsmath,graphicx,bbm,color,booktabs}
\usepackage{amsopn}

\newcommand{\vz}{\bm{z}}               

\DeclareMathOperator*{\argmax}{argmax}

\newcommand{\zf}{\mathbf{z}_{F_0}}
\newcommand{\zspk}{\mathbf{z}_{spk}}
\newcommand{\zemo}{\mathbf{z}_{emo}}
\newcommand{\zdur}{\mathbf{z}_{dur}}
\newcommand{\z}{\mathbf{z}}
\newcommand{\zc}{\mathbf{z}_c}
\newcommand{\x}{\mathbf{x}}

\newcommand{\xhat}{\hat{\mathbf{x}}}

\newcommand{\oursp}{\textsc{Ours}^{+}} 
\newcommand{\oursn}{\textsc{Ours}^{-}} 

\usepackage[T1]{fontenc}
\usepackage[utf8]{inputenc}
\usepackage{microtype}

\title{Textless Speech Emotion Conversion using \\ Discrete \& Decomposed Representations}

\author{Felix Kreuk$^{1,2}$\thanks{~~~Work done while Felix Kreuk was an Intern at Meta AI Research.}, Adam Polyak$^2$, Jade Copet$^2$, Eugene Kharitonov$^2$, Tu-Anh Nguyen$^2$, \\ \textbf{Morgane Rivière$^2$, Wei-Ning Hsu$^2$, Abdelrahman Mohamed$^2$, Emmanuel Dupoux$^{2,3}$, Yossi Adi$^2$} \\
        Bar-Ilan University, Ramat-Gan, Israel$^1$ \\ 
        Meta AI Research$^2$ \\ EHESS, Paris$^3$\\
        \texttt{\{felixkreuk, adiyoss\}@fb.com}}

\begin{document}
\maketitle
\begin{abstract}
\emph{Speech emotion conversion} is the task of modifying the perceived emotion of a speech utterance while preserving the lexical content and speaker identity. In this study, we cast the problem of emotion conversion as a spoken language translation task. We use a decomposition of the speech signal into discrete learned representations, consisting of phonetic-content units, prosodic features, speaker, and emotion. First, we modify the speech content by translating the phonetic-content units to a target emotion, and then predict the prosodic features based on these units. Finally, the speech waveform is generated by feeding the predicted representations into a neural vocoder. Such a paradigm allows us to go beyond spectral and parametric changes of the signal, and model non-verbal vocalizations, such as laughter insertion, yawning removal, etc. We demonstrate objectively and subjectively that the proposed method is vastly superior to current approaches and even beats text-based systems in terms of perceived emotion and audio quality. We rigorously evaluate all components of such a complex system and conclude with an extensive model analysis and ablation study to better emphasize the architectural choices, strengths and weaknesses of the proposed method. Samples are available under the following link: 
\href{https://speechbot.github.io/emotion}{[samples]}.
\end{abstract}

\section{Introduction}
\label{sec:intro}
\begin{figure}[t!]
\centering
\resizebox{0.48\textwidth}{!}{%
  \includegraphics[width=\linewidth]{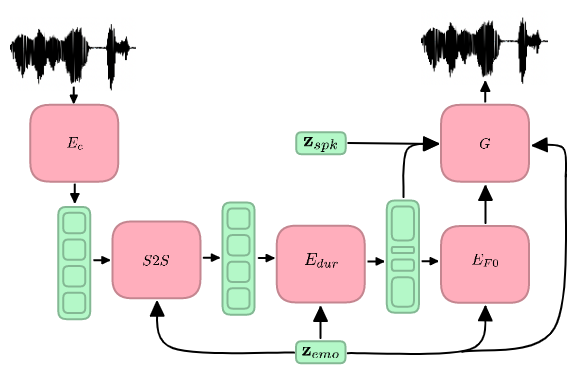} 
}
\caption{An illustration of the proposed system. 
The input signal is first encoded as a discrete sequence of content units ($E_{c}$). Next, a sequence to sequence ($S2S$) model is applied to translate between the sequences corresponding to different emotions. Then we predict the duration ($E_{dur}$) and F0 ($E_{F0}$) before feeding these signals to a vocoder ($G$). The last 4 components are conditioned by emotion ($z_{emo}$). 
We use pink to denote models and green to denote representations.}
\label{fig:arch}
\end{figure}

Generating spoken utterances and dialogue that sound natural is a fundamental requirement to improve human-computer interaction~\cite{tits2019theory}. One of the main roadblock in improving naturalness in speech generation is the modeling of expressive and emotional states. The difficulty is that emotion is a phenomenon affecting all linguistic levels simultaneously: when one goes from a happy to an angry state, one may use different vocabulary, insert non-verbal vocalizations (cries, grunts, etc), modify prosody (intonation and rhythm), and change voice quality due to stress. Vice versa each of the levels contributes to the perception of the emotional state of the speaker, where the non verbal aspects can often override the lexical content~\cite{mehrabian1967decoding}.

Existing emotion generation or emotion conversion techniques have a hard time producing convincing results because they only manage to tackle a subset of these levels. 
In a nutshell, signal-based approaches are mainly focused on manipulating parameters of the speech signal~\cite{inanoglu2009data, aihara2012gmm, gao2018nonparallel, robinson2019sequence, polyak2020unsupervised, rizos2020stargan, zhou2020transforming, zhou2020converting, zhou2021limited, zhou2021seen, zhou2021vaw}, and can only address changes at the level of voice and prosody. In contrast, text based approaches ~\cite{skerry2018towards, fastspeech2, kim2021expressive, sorin2020principal, hono2020hierarchical,tan2021survey, polyak2019tts} can generate expressive speech, but struggle with nonverbal vocalizations because they are typically not annotated in speech corpora.

In this work we focus on the task of speech emotion conversion under the parallel dataset setting, modifying the perceived emotion of a speech utterance while preserving the speaker identity and the lexical content~\cite{zhou2021emotional}.

The proposed pipeline is comprised of four main blocks: speech tokenizer, content translation model, prosody prediction model, and a neural vocoder. We start by extracting discrete representation of the speech signal. We translate these representation to a target emotion (e.g., removing laughter, inserting yawning) while preserving the lexical content. Then, we predict prosodic features based on the translated representations. A neural vocoder synthesizes the speech waveform from the translated phonetic-content, predicted prosody, speaker label and target emotion label. The overall system is depicted in Figure~\ref{fig:arch}.

Our contribution is twofold:
We propose a novel textless approach by casting the task of speech emotion conversion as a \emph{spoken language translation} problem. We demonstrate how such a paradigm can be used to model expressive non-verbal communication cues as well as generating high-quality speech samples. Finally, we demonstrate for the first time the coverage of all levels of expressive speech modeling simultaneously.
Results show that the proposed method is vastly superior to emotion conversion techniques based on signal only, and also beats text-based approaches in terms of generation quality and perceived emotion. 

\section{Related work}
\label{sec:related}
\paragraph{Speech emotion conversion.} 
There is an increasing number of studies for speech emotion conversion recently~\cite{zhou2021emotional}. Early studies on speech emotional conversion include \ac{GMM} technique~\cite{aihara2012gmm}, sparse representation technique~\cite{aihara2014exemplar}, or an incorporated framework of \ac{HMM}, \ac{GMM} and fundamental frequency segment selection method~\cite{inanoglu2009data}. Recently, speech emotion conversion has seen a great leap in performance following the advances in deep nets. This include using multi-layer perceptron~\cite{luo2016emotional, lorenzo2018investigating}, highway networks~\cite{shankar2019automated}, deep bi-directional LSTM networks~\cite{ming2016deep}, and sequence-to-sequence model~\cite{robinson2019sequence}. When considering non-parallel data, new techniques have been proposed to learn the translation between emotional domains with CycleGAN~\cite{zhou2020transforming} and StarGAN~\cite{rizos2020stargan}, to disentangle the emotional elements from speech with auto-encoders~\cite{cao2020nonparallel,zhou2021vaw}, and to leverage text-to-speech ~\cite{kim2020emotional} or \ac{ASR}~\cite{liu2020multi}. Studies have also revealed that the emotions can be expressed through universal principles that are shared across different individuals and cultures~\cite{ekman1992argument, manokara2021people}. This motivates research on multispeaker~\cite{shankar2019multi, shankar2020multi}, and speaker-independent emotion conversion~\cite{zhou2020converting, choi2021sequence}.

\paragraph{Unsupervised speech representation learning} aims to distill features useful for downstream tasks, such as phone discrimination~\cite{kharitonov2020data,schneider2019wav2vec} and semantic prediction~\cite{lai2021semi,wu2020self}, by constructing pretext tasks that can exploit large quantities of unlabeled speech. Pretext tasks in the literature can be roughly divided into two categories: reconstruction and prediction. 
Reconstruction is often implemented in the form of auto-encoding~\cite{van2017neural}, where speech is first encoded into a low-dimensional space, and then decoded back to speech. Various constraints can be imposed on the encoded space, such as temporal smoothness~\cite{ebbers2017hidden,glarner2018full,khurana2020convolutional}, discreteness~\cite{ondel2016variational,van2017neural}, presence of hierarchy~\cite{hsu2017unsupervised}, and information bottlenecks for speech representation decomposition~ \cite{qian2020unsupervised,qian2021global}.
Prediction-based approaches task a model with predicting information of unseen speech based on its context. Examples of information include spectrograms~\cite{wang2020unsupervised,chi2020audio,chung2020improved}, cluster indices~\cite{hsu2020hubert}, derived signal processing features~\cite{ravanelli2020multi}, and binary labels of whether a candidate is the target unseen spectrogram~\cite{schneider2019wav2vec,kharitonov2020data,baevski2020wav2vec}.

\paragraph{Spoken Language Modeling.} The closest related work to ours concerns with modeling spoken language without textual supervision. \citet{gslm} suggested utilizing recent success of self-supervised speech representation learning for discovering discrete units and modeling them. Then, conditional and unconditional speech generation can be achieved by sampling unit sequences from a unit-Language-Model, and synthesizing them to speech using a units-to-Speech  model~\cite{gslm}. Moreover, the authors proposed a set of evaluation functions to assess the quality of the overall system. Next,~\citet{gslm2} evaluated the robustness and disentanglement properties of several Speech-to-units models and demonstrated the ability to perform voice conversion, and lightweight speech codec. Lastly,~\citet{gslm3} proposed a multi-stream transformer model to jointly process ''pseudo-text`` units and prosodic features (i.e., duration and F0). This was shown beneficial in improving content modeling capabilities. In this work, we leverage a similar approach and speech representation scheme but for speech emotion conversation via translation.
\vspace{-0.2cm}
\section{Model}
\label{sec:model}
\vspace{-0.1cm}
As emotion manifests itself in multiple aspects of spoken language, to optimally convert emotion one needs to consider all aspects in the conversion process. For example, emotion can be expressed via a prosodic features (high pitch, slow speaking rate, etc.), speaking style (yelling, whispering, etc.), and non-verbal vocalizations (laughing, yawning, etc.).

We suggest to use a decomposed representation of the speech signal to synthesize speech in the target emotion. We consider four components in the decomposition: phonetic-content, prosodic features (i.e., F0 and duration), speaker identity, and emotion-label, denoted by $\zc, (\zdur, \zf), \zspk, \zemo$ respectively. 

Specifically, we propose the following cascaded pipeline: (i) extract $\zc$ from the raw waveform using a \ac{SSL} model; (ii) translate non-verbal vocalizations in $\zc$ while preserving the lexical content (e.g., when converting from amused to sleepy, we remove laughter and insert yawning); (iii) predict the prosodic features of the target emotion based on the translated content; (iv) synthesize the speech from the translated content, predicted prosody, target speaker identity and target emotion-label. See Figure~\ref{fig:arch} for a visual description of the method.

\subsection{Speech Input Representation}
\paragraph{Phonetic-content representation.}
To represent speech phonetic-content we extract a discrete representation of the audio signal using a pre-trained \ac{SSL} model, namely HuBERT. We use a \ac{SSL} representation for phonetic-content in order to capture non-verbal vocalizations (unlike text where they are often not annotated). We discretize this representation for better modeling and sampling (as opposed to regressing on continuous variables). This paradigm allows us to benefit from all recent advances in \ac{NLP}. We chose HuBERT for the phonetic-content units as it was shown to better disentangle between speech content and both speaker and prosody compared to other \ac{SSL}-based models~\cite{gslm2}.

Denote the domain of audio samples by $\mathcal{X} \subset \mathbb{R}$. The representation for an audio waveform is therefore a sequence of samples $\x = (x^1, \dots, x^T)$, where each $x^i \in \mathcal{X}$ for all $1 \leq t \leq T$. The content encoder $E_c$ is a HuBERT model \cite{hsu2020hubert} pre-trained on the LibriSpeech corpus~\cite{Panayotov2015}. HuBERT is a self-supervised model trained on the task of masked prediction of continuous audio signals, similarly to BERT~\cite{bert}. During training, the targets are obtained via clustering of MFCCs features or learned representations from earlier iterations. The input to the content encoder $E_c$ is an audio waveform $\x$, and the output is a spectral representation sampled at a lower frequency $\z' = (z_c^{'1}, \dots, z_c^{'L})$ where $L < T$. Since HuBERT outputs continuous representations, an additional k-means step is needed in order to quantize these representations into a discrete unit sequence denoted by $\zc = (z_c^1, \dots, z_c^L)$ where $z_c^i \in \{1, \dots, K\}$ and $K$ is the size of the vocabulary. For the rest of the paper, we refer to these discrete representations as ``units''. We extracted representations from the 9$^{th}$ layer of HuBERT model and set $K=200$. Following~\citet{gslm}, repeated units were omitted (e.g., $0,0,0,1,1,2 \rightarrow 0,1,2$). We denote such sequences by ``deduped''. 

\paragraph{Speaker representation.}
Our goal is to convert speech emotion while keeping the speaker identity fixed. To that end, we construct a speaker-representation $\zspk$, and include it as an additional conditioner during the waveform synthesis phase. To learn $\zspk$ we optimize the parameters of a fixed size look-up-table. Although such modeling limits our ability to generalize to new and unseen speakers, it produces higher quality generations~\cite{gslm2}. 
We additionally experimented with representing $\zspk$ as a \emph{d-vector} using a method similar to~\citet{heigold2016end}. However, we observed that such approach keeps source emotion prosodic features, resulting in inferior disentanglement during the waveform synthesis phase.

\paragraph{Emotion-label representation.} 
We represent the emotion-label using a categorical variable represented by a 1-hot vector. We observed that this component controls for timbre characteristics of the generated speech signal (e.g., roughness, smoothness, etc.) during the synthesis phase.

\subsection{Speech Emotion Conversion}
\label{sec:speech_modeling}
Using the above representations we propose to synthesize the speech signal in the target emotion. We use a translation model to convert between phonetic-content units of a source emotion to phonetic-content units of the target emotion. This serves as a learnable insertion/deletion/substitution mechanism for non-verbal vocalizations, while preserving the lexical content (e.g., removing yawning while preserving the verbal content). Next, we predict the prosodic features (duration and F0) based on the translated phonetic-content units and target emotion-label, and inflate the sequence according to the predicted durations. This will later be used as a conditioning for the waveform synthesis phase. 

\begin{figure}[t!]
\centering
\resizebox{0.48\textwidth}{!}{%
  \includegraphics[width=\linewidth]{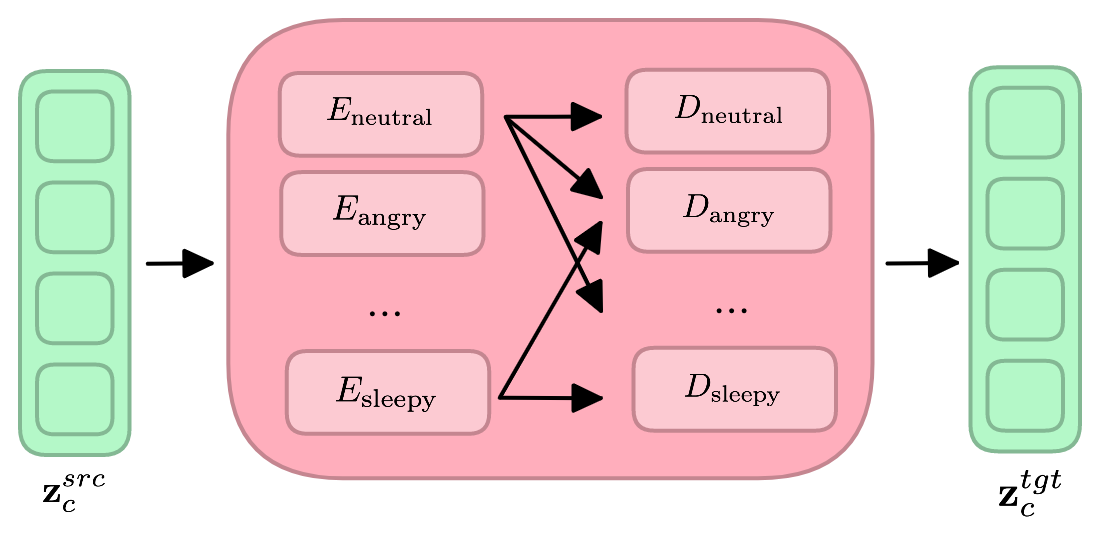}  
}
\caption{An illustration of the sequence-to-sequence emotion translation component $E_{s2s}$. Here different encoders and decoders are used for each emotions, but we also tested shared architectures in Appendix \ref{sup:add_res}. }
\label{fig:arch_s2s}
\end{figure}

\paragraph{Unit translation.}
To translate the speech content units, we use a sequence-to-sequence Transformer model~\cite{trans} denoted by $E_{s2s}$. The input to $E_{s2s}$ is a sequence of phonetic-content units representing a speech utterance in the source emotion $\zc^{src}$.
The model is trained to output a sequence of phonetic-content units $\zc^{tgt}$ containing the same lexical content with the addition/deletion/substitution of speech cues related to emotion expression (e.g., inserting laughter units).
The optimization minimizes the cross-entropy (CE) loss between the predicted units $\hat{\zc}=E_{s2s}(\zc^{src}, \zemo)$ and ground-truth units for each location in the sequence,
\begin{align}
\label{eq:s2s}
    L_{s2s} = \sum_{i=1}^{L} CE(E_{s2s}( {\zc^{src}, \zemo)}_i, {\zc^{tgt}}_i).
\end{align}
In our experiments, we observed that directly optimizing the above model for translation captures emotion transfer, but fails to maintain the same lexical content, producing expressive yet unintelligible speech utterances (e.g., the model adds laughter but corrupts the sentence by removing needed syllables). To mitigate this, we pre-train the translation model on the task of language denoising auto-encoder similarly to BART \cite{bart}. To better support translation between all emotions we use a dedicated encoder and decoder for each emotion (see Figure~\ref{fig:arch_s2s}). We additionally evaluated a shared-encoder and shared-decoder model, in which we condition on the target emotion, as well as a model where only the encoder is shared. We evaluated all approaches, results are summarized in Table~\ref{tab:arch} Appendix \ref{sup:add_res}.

\paragraph{Prosody prediction.} 
Next, using the translated phonetic-content unit sequence we predict the prosodic features corresponding to the target emotion. We consider the prosodic representation as a tuple of content unit durations and F0. 

We start by describing the duration prediction process. Due to working on deduped sequences, we first need to predict the duration of each phonetic-content unit. We follow a similar approach to the one proposed in~\cite{fastspeech2} and use a \ac{CNN} to learn the mapping between content units to durations. We denote this model by $E_{dur}$.
During training of $E_{dur}$, we input the deduped phonetic-content units $\zc$ and use the ground-truth phonetic-content unit durations as supervision. We minimize the \ac{MSE} between the network's output and the target durations. We also evaluated n-gram-based duration prediction models. The n-gram models were trained by counting the mean frequency $\mu$ and the standard deviation $\sigma$ of each n-gram in the training set. During inference, we predict the duration of each n-gram by sampling from $\mathcal{N}(\mu, \sigma)$. For unseen n-grams we back-off to a smaller n-gram model. Results can be found on Table~\ref{tab:dur} in the Appendix~\ref{sup:add_res}.

We now turn to describe the F0 prediction process. We use a F0 estimation model to predict the pitch from a sequence of phonetic-content units $\zc$. Our model, denoted by $E_{F0}$, is a \ac{CNN}, followed by a linear layer projecting the output to $\mathbb{R}^{d}$. The final activation layer is set to be a sigmoid such that the network outputs a vector in $[0,1]^{d}$. We extract the F0 using the YAAPT~\cite{yaapt} algorithm to serve as targets during training. Next, we normalize the F0 values using the mean and standard deviation per speaker. We follow a similar approach to~\citet{kim2018crepe} and discretize the range of F0 values into $d$ bins represented by one-hot encodings. Next, we apply Gaussian-blur on these encodings to get the final supervision targets denoted by $\vz_{F0}=(z_{F0}^1, \dots, z_{F0}^{T'})$ where each $z_{F0}^i \in [0,1]^d$, and $d=50$. Formally, we minimize the binary-cross-entropy (BCE) for each coordinate of the target and the network output,
\begin{align}
    L_{F0} = \sum_{i=1}^{d} BCE(E_{F0}(\zc, \zemo^{tgt})_i, \vz_{F0}^i).
\end{align}

During inference multiple frequency bins can be activated to a different extent, similarly to \citet{kim2018crepe} we output the F0 value corresponding to the weighted-average of the activated bins. This modeling allows for a better output range when converting bins back to F0 values, as opposed to a single representative F0 value per bin. For F0-to-bin conversion, we use an adaptive binning strategy such that the probability mass of each bin is the same, similarly to~\citet{gslm3}. 
Results are summarized on Table~\ref{tab:f0_log} Appendix \ref{sup:add_res} together with additional comparisons to log-F0 estimation models, uniform binning, and $\argmax$ decoding.

Notice, the mapping between discrete unit sequences and prosodic feature (F0 and durations) is one-to-many, as it depends on the target emotion. Hence, we additionally condition both $E_{dur}$ and $E_{F0}$ on the target emotion denoted by $\zemo^{tgt}$. Both model were trained independently.

\paragraph{Speech synthesis.}
We follow~\citet{gslm2}, and use a variation of the HiFi-GAN neural vocoder~\cite{kong2020hifi}. The architecture of HiFi-GAN consists of a generator $G$ and a set of discriminators $D$. We adapt the generator component to take as input a sequence of predicted phonetic-content units inflated using the predicted durations, predicted F0, target speaker-embedding, and a target emotion-label. The above features are concatenated along the temporal axis and fed into a sequence of convolutional layers that output a 1-dimensional signal. The sample rates of unit sequence and F0 are matched by means of linear interpolation, while the speaker-embedding and emotion-label are replicated.

The discriminators are comprised of two sets: Multi-Scale Discriminators (MSD) and Multi-Period Discriminators (MPD). The first type operates on different sizes of a sliding window over the signal (2, 4), while the latter samples the signal at different rates (2,3,5,7,11). Overall, each discriminator $D_i$ is trained by minimizing the following loss function,
$ L_{adv}(D_i, G) = \sum_{\x} || 1 - D_i(\xhat) ||^2_2 $ and 
$ L_D(D_i, G) = \sum_{\x} || 1 - D_i(\x) ||^2_2 + || D_i(\xhat) ||^2_2 $, 
where $\xhat = G(\hat{\zc}, E_{F0}(\hat{\zc}, \zemo), \zspk, \zemo)$ is the time-domain signal reconstructed from the decomposed representation. There are two additional loss terms used for optimizing $G$. The first is a Mean-Absolute-Error (MAE) reconstruction loss in the log-mel frequency domain $L_{recon}(G) = \sum_{\x} || \phi(\x) - \phi(\xhat) ||_1$.
where $\phi$ is the spectral operator computing the Mel-spectrogram. The second loss term is a feature matching loss, which penalizes for large discrepancies in the intermediate discriminator representations, $L_{fm}(D_i, G) = \sum_{\x}\sum^{R}_{j=1} || \xi_j(\x) - \xi_j(\xhat) ||_1$
where $\xi_j$ is the operator that extracts the intermediate representation of the $j$th layer of discriminator $D_i$ with $R$ layers.
The overall objective for optimizing the system is:
\begin{align}
    L_G(D,G) &= \bigg[ \sum_{i=1}^J L_{adv}(D_i, G) + \lambda_{fm} L_{fm}(D_i, G)\bigg] \notag \\ 
    &+ \lambda_{r} L_{recon}(G), \notag 
\end{align}
and $L_D(D,G){=}\sum_{i=1}^J L_D(D_i, G)$, where $\lambda_{fm}=2$ and $\lambda_r=45$. 
\section{Experimental Setup}
\label{sec:exp}
We use the Emotional Voices Database (EmoV) for training and evaluating our model.
EmoV consists of 7000 speech utterances based on transcripts from the CMU Arctic Database \cite{kominek2004cmu}.
Each transcript was recorded in multiple acted emotions (neutral, amused, angry, sleepy, disgusted) by multiple native speakers (two males speakers and two females speakers). 

This allows us to create a dataset of utterance pairs for the task of translation. Specifically, we create pairs of utterances that are based on the same transcript but are recorded with different acted emotions. Due to the small size of this dataset ($\sim$9 hours), we further augment it by creating additional parallel pairs from different speakers. Overall, the size of the entire dataset is 78,324 pairs. We split the data to train/validation/test sets with a ratio of 90/5/5 such that there is no overlap of utterances between the sets. In our experiments, splitting randomly (e.g., overlapping transcripts) led to a memorization of the utterances and failed to generalize to unseen data. 
We explore two data regimes for pre-training stage: (1) large scale pre-training on a mix of Librispeech, Blizzard2013~\cite{blizzard} and EmoV~\cite{emov}. Denoted by $\oursp$; (2) small scale setup in which we pre-train on VCTK~\cite{yamagishi2019cstr} and EmoV. We explore the latter for a fair comparison with the evaluated baselines. We denote this setup by $\oursn$.
A full description of architecture hyperparameters and pre-training details can be found in the Appendix~\ref{sup:imp_det}.

\subsection{Baselines}
\label{sec:baselines}
We compare the proposed method to a textless speech emotion conversion method, VAW-GAN~\cite{zhou2020converting}, as well as a \ac{SOTA} text-based emotional voice conversion model, Seq2seq-EVC \cite{zhou2021limited}. We also evaluate an expressive \ac{TTS} system based on Tacotron2~\cite{shen2018natural}. \footnote{We consider text-based systems as ones using textual annotations during training.} 

For the text-based approach we use the Tacotron2 and Seq2seq-EVC models. The input to Tacotron2 is the ground-truth text representing the speech content. We modify the Tacotron2 architecture by adding a Global-Style-Token \cite{expressive_taco} to control for the target emotion. The inputs to Seq2seq-EVC are the ground-truth phonemes coupled with the source speech utterance. The output of both systems is the Mel-spectrogram of the speech utterance in the target emotion. To reconstruct the time-domain signal we use the HiFi-GAN vocoder.

All baselines were trained and evaluated on the EmoV dataset. Seq2seq-EVC and Tacotron2 were first pre-trained on VCTK.

\subsection{Evaluation}

\paragraph{Subjective Evaluation.}
Recall, our goal is to perform speech emotion conversion. To that end, we propose a new subjective metric called \ac{eMOC}. In an \ac{eMOC} study, a human rater is presented with a speech utterance and a set of emotion categories. The rater is instructed to select the emotion that best fits the speech utterance. The \ac{eMOC} score is the percentage of raters that selected the target emotion given a speech recording. The final score is averaged over all raters and utterances in the study. Additionally, we measure the perceived audio-quality using the \ac{MOS}. 
When comparing MOS and eMOC against baselines, we report results for emotion conversion from Neutral as all baselines are constrained to Neutral as the source emotion. With provide additional results of our system in the any-to-any conversion setting.
The CrowdMOS package~\cite{ribeiro2011crowdmos} was used in all subjective experiments with the recommended recipes for outliers removal. Participants were recruited using a crowd-sourcing platform.

\paragraph{ASR Based.}
Our goal is to perform speech emotion conversion while preserving the lexical content of the speech signal. However, to compare different translation models with different vocabularies, one cannot simply use metrics such as BLEU, hence, we report \ac{WER} and \ac{PER} metrics extracted using a pre-trained \ac{SOTA} \ac{ASR} system~\footnote{We use a \textsc{base} wav2vec 2.0 phoneme detection model trained on LibriSpeech-960h with CTC loss from scratch.}. As \ac{ASR} models suffer from performance degradation when evaluated on expressive speech, \ac{WER} and \ac{PER} metrics are reported on emotions converted to neutral: \{amused, angry, sleepy, disgusted\}-neutral.
\begin{table}[t!]
    \centering
    \resizebox{0.85\columnwidth}{!}{%
    \begin{tabular}{c|c|c|c|c}
        \toprule
        \# units & layer no. & pre-trained  & WER & PER \\
        \toprule
         100    & 6 & \ding{55} & 88.85 & 72.17 \\ \hline
         100    & 6 & \ding{51} & 36.01 & 29.67 \\ \hline
         100    & 9 & \ding{55} & 83.72 & 66.63 \\ \hline
         100    & 9 & \ding{51} & 31.69 & 27.90 \\ \hline
         200    & 6 & \ding{55} & 91.06 & 77.14 \\ \hline
         200    & 6 & \ding{51} & 31.97 & 27.49 \\ \hline
         200    & 9 & \ding{55} & 84.92 & 71.45 \\ \hline
         200    & 9 & \ding{51} & \bf 23.92 & \bf 25.95 \\
         \bottomrule
    \end{tabular}}
    \caption{Evaluation of different token extraction configurations and the effect of pre-training the translation model. \# units denotes the vocabulary size, layer no. denotes the layer index used in HuBERT.}
    \vskip -1em
    \label{tab:abl}
\end{table}

\section{Results}
\label{sec:results}
We first tune independently the different components of our system using objective metrics: the content units extraction configuration, and the prosodic modeling modules (F0 and duration estimators). Next we conduct a subjective evaluation for our best system in terms of audio-quality (\ac{MOS}) and perceived emotions (\ac{eMOC}) and compare the proposed method against the baselines (Section~\ref{sec:baselines}). Finally, we run an ablation study where we measure the impact of each component on the perceived emotion. 

\subsection{Model Tuning}
\citet{yang2021superb} found that intermediate HuBERT representations obtained from different layers have an impact on the downstream task at hand. Specifically, when considering spoken language modeling, \citet{gslm} used the 6$^{th}$ layer, while \citet{gslm3} used layer 9. Additionally, \citet{gslm} showed that the number of units ($k$) also has an impact on the overall performance of the system. 
To better understand the effect of these architectural configurations in our setting, we experimented with extracting HuBERT features from layers 6 and 9, using 100 and 200 clusters for the k-means post-processing step. Additionally, we measure the impact that pre-training the translation model has on performance. As we compare models with different vocabulary sizes, we cannot use metrics such as BLEU, hence, we report \ac{WER} and \ac{PER}. 

For emotion conversion, we find that using the 9$^{th}$ HuBERT layer and 200 tokens performs best. An evaluation of different model design architectures can be found in Table~\ref{tab:abl}.

\begin{figure}[t!]
\centering
\begin{subfigure}{0.45\textwidth}
  \centering
  \includegraphics[width=\linewidth]{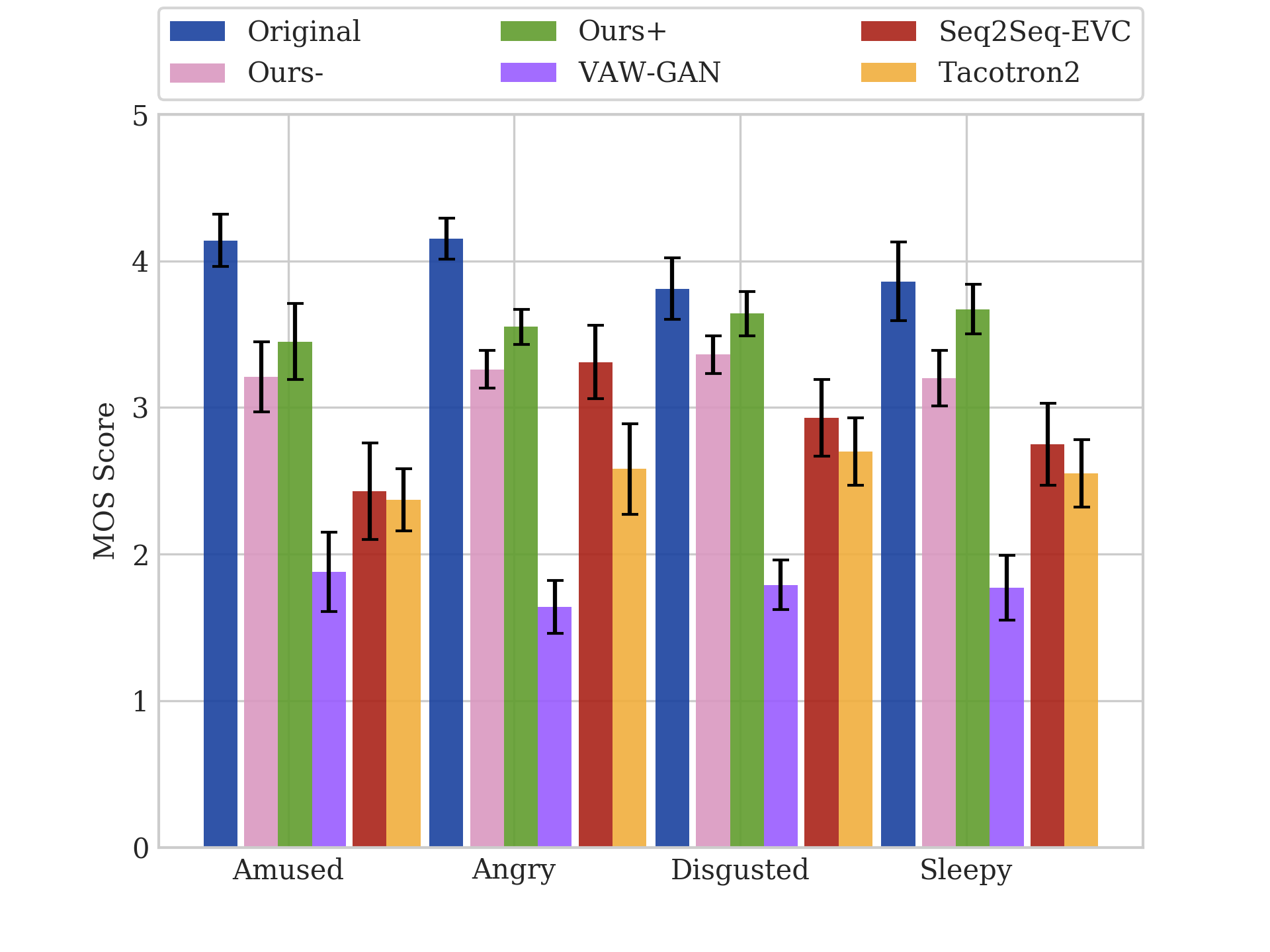}  
  \caption{}
  \label{fig:ovl_mos}
\end{subfigure}
\begin{subfigure}{0.45\textwidth}
  \centering
  \includegraphics[width=\linewidth]{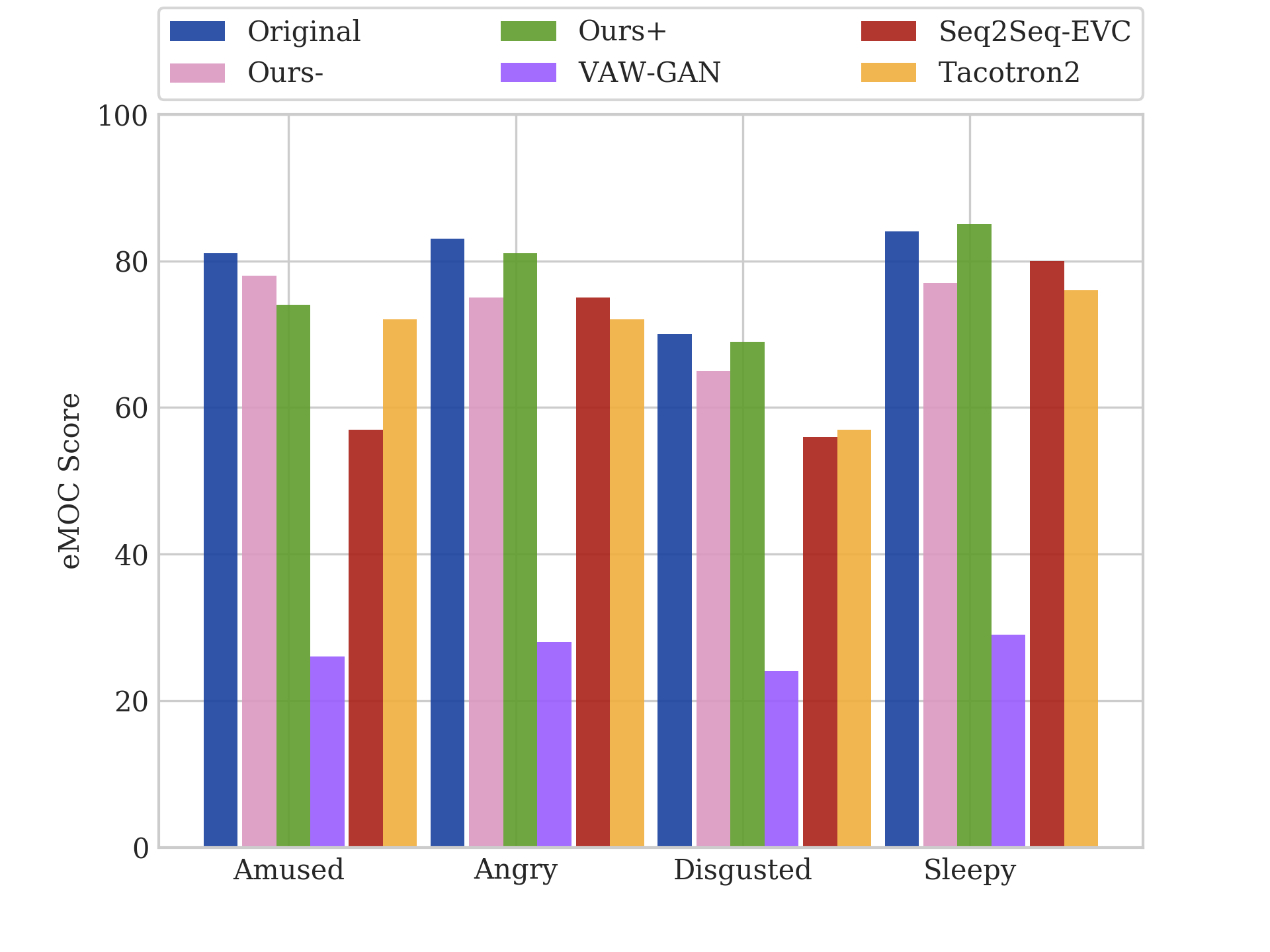}  
  \caption{}
  \label{fig:ovl_emos}
\end{subfigure}
\caption{Sub-figure (a) reports the \ac{MOS}, measuring the perceived audio-quality. We report mean scores with a Confidence Interval of 95\%. Sub-figure (b) reports the \ac{eMOC} score, measuring the perceived emotion. We report mean scores for each emotion (chance level: 25\%). $\oursp$ was pre-trained on Librispeech, Blizzard2013 and EmoV, and $\oursn$ was pre-trained on VCTK and EmoV.}
\label{fig:mos}
\end{figure}

\subsection{Subjective Evaluation}
Figure~\ref{fig:mos} depicts the \ac{MOS}, and \ac{eMOC} scores for the proposed method and the evaluated baselines. For a fair comparison against baselines, all emotion  were  converted from Neutral. Results suggest that both $\oursn$ and $\oursp$ surpass the baselines in terms of both \ac{MOS} and \ac{eMOC}, with the exception \ac{MOS} score for Angry and \ac{eMOC} for Sleepy where $\oursn$ is comparable to Seq2seq-EVC. 

While Tacotron2 and Seq2seq-EVC succeed in conveying the target emotion, they produce less natural expressive speech utterances, which is reflected in lower MOS. 
Unlike our method, both Tacotron2 and Seq2seq-EVC are text-based systems hence they attempt to learn an alignment (an attention map) between text inputs and audio targets. This task is particularly challenging as non-verbal cues (e.g., laughter, breathing) are not annotated in the text inputs. We hypothesize that this misalignment leads to less natural expressive speech production.
We provide additional \ac{eMOC} results for the proposed method, evaluated baselines, and ground truth recordings in the Appendix~\ref{sup:add_res}.

\begin{figure}[t!]
\centering
\begin{subfigure}{0.45\textwidth}
  \centering
  \includegraphics[width=\linewidth]{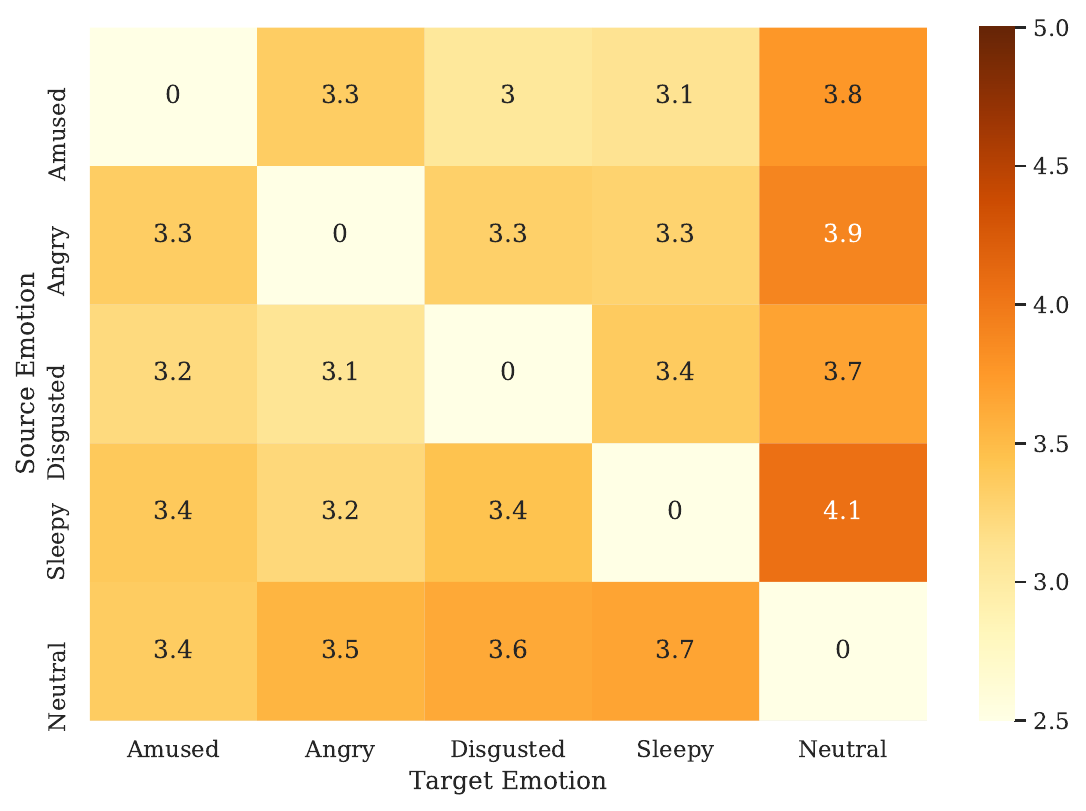}  
  \caption{MOS}
  \label{fig:all2all_mos}
\end{subfigure}
\begin{subfigure}{0.45\textwidth}
  \centering
  \includegraphics[width=\linewidth]{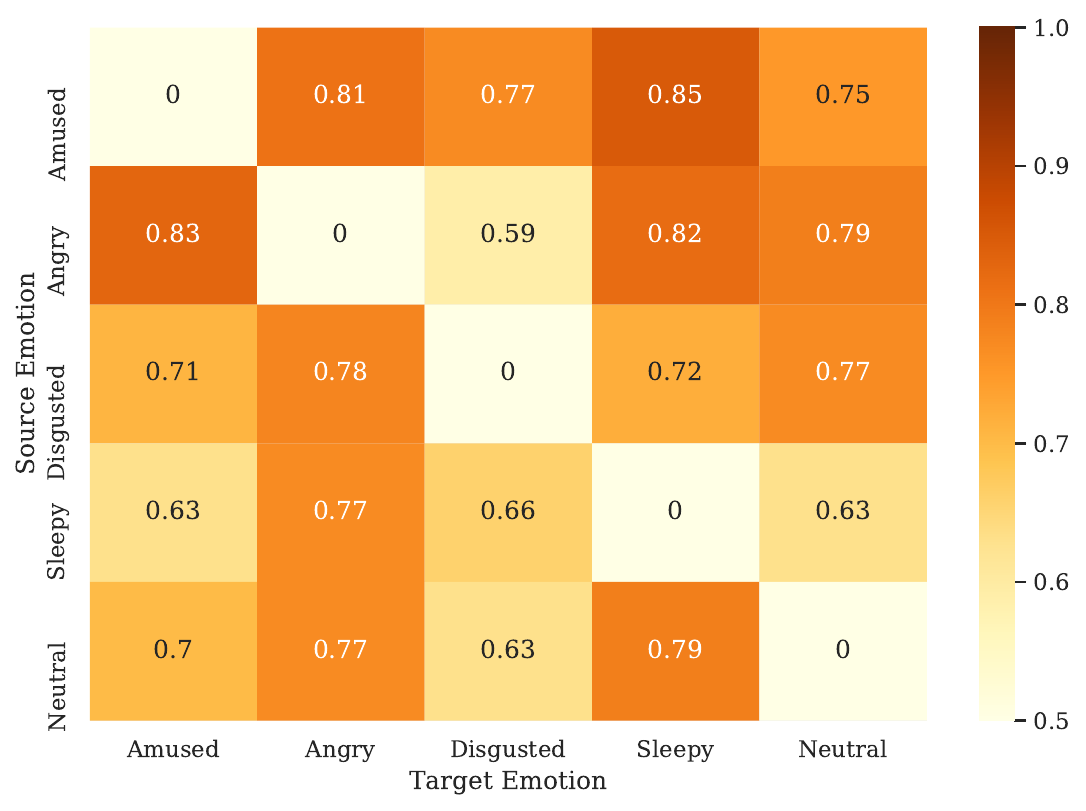}  
  \caption{eMOC}
  \label{fig:all2all_emos}
\end{subfigure}
\caption{MOS and eMOC results for many-to-many emotion conversion. We report MOS score for each pair of source and target emotions. For eMOC we report percentage of raters that chose the target emotion.}
\label{fig:all2all}
\end{figure}

The evaluated baselines considered only Neutral as the source emotion. As the proposed method supports many-to-many~\footnote{from a set of pre-defined emotions} emotion voice conversion, we additionally report MOS and eMOC results when converting from any emotion to any emotion. 
Figure~\ref{fig:all2all} summarizes the MOS and eMOC scores for the many-to-many setting.
We additionally provide the full eMOC results for the many-to-many setting in the Table~\ref{sup:emov_all2all_full} of the Appendix~\ref{sup:add_res}.
In terms of MOS, results suggest that converting to Neutral and from Neutral produced the most natural sounding utterances.
In terms of eMOC, having Amused as the source emotion surprisingly yielded the highest overall accuracy, with $\sim$80\% of conversions successful (average of first row in Figure~\ref{fig:all2all_emos}). On the other hand, the Sleepy emotion was the hardest to convert, with 12\% of raters still choosing Sleepy after conversion. The pair of emotions that were hardest to distinguish are Disgusted and Neutral.

\subsection{Ablation}
Recall, we use a decomposed speech representation consisting of four feature sets. 
Here, we gauge the effect of each feature by gradually adding different components and evaluating their impact on the \ac{eMOC} metric. Specifically, we start by evaluating the source features and replacing the emotion-token. Then, we predict the unit durations and F0 for the target emotion. Lastly, the full effect of our method is achieved by incorporating the unit translation model. For reference, we report results for the original recording, and resynthesized one (i.e., using source features only) and the target recording. Results are summarized in Table~\ref{tab:abl_comp}. 

Results suggest that the proposed method is comparable to ground-truth recordings in terms of the perceived emotion. Interestingly, for the Sleepy emotion, modifying the timbre using the target emotion-token and adjusting the unit durations is enough to reach $70.31\%$, while the rest of the emotions require further processing. For Amused, Angry and Disgusted modifying the F0 reaches performance of $5\%$ below the ground-truth recordings. When applying the entire pipeline results are on par with the ground-truth. 


\subsection{Out-of-distribution samples}
Due to the size of the EmoV dataset (7000 samples overall) the number of unique utterance is small. As a result, the model can memorize the utterances and might fail to generalize to unseen sentences. Hence, we experimented with converting out-of-domain recordings. To that end, we input our system with recordings from the LibriSpeech dataset. As LibriSpeech consists of non-expressive samples, we treat them as ``neutral''. We convert these samples to different emotions and evaluate the performance. For evaluation we randomly sampled 20 utterances converted to each emotion. The proposed method reaches an average \ac{eMOC} score of $82.25\% \pm7.32$ and an average \ac{MOS} score of $3.69 \pm0.26$ across all four emotions (amused, angry, disgusted, sleepy). When considering the lexical content, the \ac{WER} between the ground-truth text and \ac{ASR} based transcriptions of the generated audio is 27.92. These results are similar to the ones reported for the EmoV dataset.

\begin{table}[t!]
    \centering
    \resizebox{1.0\columnwidth}{!}{%
        \begin{tabular}{c|c|c|c|c|c|c|c}
        \toprule
         \multirow{2}{*}{Tokens} & \multirow{2}{*}{Emotion} & \multirow{2}{*}{F0} & \multirow{2}{*}{Duration} & \multicolumn{4}{c}{\ac{eMOC}} \\
         & & & & Amused & Angry & Disgusted & Sleepy\\
        \toprule
        \multicolumn{4}{c|}{Original -- Neutral}        & 20.11 & 21.66 & 7.59 & 17.34  \\ \midrule \midrule
        src  & src   & src  & src                       & 20.24 & 18.62 & 8.37 & 17.72   \\ \midrule
        src  & tgt   & src  & src                       & 27.66 & 25.76 & 10.83 & 25.19  \\ \midrule
        src  & tgt   & src  & pred                       & 30.89 & 35.71 & 32.11 & 70.31 \\ \midrule
        src  & tgt   & pred & pred                      & 79.12 & 86.11 & 69.21 & 79.12  \\ \midrule
        pred & tgt   & pred & pred                      & 85.16 & 90.61 & 75.89 & 84.23  \\ \midrule\midrule
        \multicolumn{4}{c|}{Original -- Emotion}        & 86.31 & 89.92 & 76.18 & 88.01  \\ 
         \bottomrule
    \end{tabular}}
    \caption{Effect of components in our system on perceived emotion. First row denotes the ground-truth neutral recordings while last row denotes the ground-truth emotional recording. ``src'', ``tgt'', and ``pred'' denote features extracted from source speech, target speech, and predicted by our system respectively.}
    \label{tab:abl_comp}
\end{table}
\section{Conclusion}
\label{sec:conclusion}
We presented a novel textless method for speech emotion conversion using decomposed and discrete representations. Using such representations we suggested casting speech emotion conversion as a spoken language translation problem. We learn a mapping to translate between discrete speech units from one emotion to another. Results suggest the proposed method outperforms the evaluated baselines by a great margin. We demonstrated how the proposed system is able to model expressive non-verbal vocalizations as well as generate high-quality expressive speech. We conclude with an ablation study and analysis of the different components composing our system.

This study serves as the foundation for improving speech emotion conversion and building general textless expressive speech generation models. For future work, we would like to design and evaluate an end-to-end system where we jointly model content and prosody, as well as extending the current work to the non-parallel dataset setting. 

\bibliography{refs}
\bibliographystyle{acl_natbib}

\appendix
\section{Appendix}
\subsection{Implementation Details}
\label{sup:imp_det}
We use the sequence-to-sequence Transformer model as implemented in \texttt{fairseq}~\cite{Ott2019}. The model contains 3 layers for both encoder and decoder modules, 4 attention heads, embedding size of 512, FFN size of 512, and dropout probability of 0.1.

For pre-training, we explore two settings: (1) large scale -- using a mix of LibriSpeech, Blizzard2013 and EmoV, denoted by $\oursp$; and (2) small scale -- using a mix of VCTK and EmoV, denoted by $\oursn$. We stop training after 3M update steps. We use the following input augmentations: \emph{Infilling} using $\lambda=3.5$ for the Poisson distribution, \emph{Token masking} with probability of 0.3, \emph{Random masking} with probability of 0.1, and \emph{Sentence permutation}. Finally, we fine-tune this model on the task of translation using paired utterances of different emotions from EmoV and early-stop using the loss from Equation~\ref{eq:s2s}. 

Our F0 prediction model $E_{F0}$ consists of six 1-D convolutional layers, where the number of kernels per layer is 256 and the respective kernel size is 5. For non-linearity we used the ReLU activation function followed by layer-norm and dropout ($p=0.1$).

Our duration prediction model $E_{dur}$ consists of two convolutional layers, where the number of kernels per layer is 256 and the respective kernel size is 3. For non-linearity we used the ReLU activation function followed by layer-norm and dropout ($p=0.5$).

\paragraph{Computational Resources.} All experiments done in this study are conducted using 8 NVIDIA V100 GPUs with 32GB memory each. 

\subsection{Additional Results}
\label{sup:add_res}

\paragraph{Model Architecture}
We evaluated three weight-sharing schemes for this model: (i) all emotions use the same encoder and decoder components of the Transformer architecture. In this case, we condition the model on the target emotion. This can be done by prepending a special target emotion token at the beginning of the decoding procedure. We denote this approach by ``share-all''. (ii) All emotions share the same encoder but have separate dedicated decoders. In this case, no target emotion conditioning is needed. We denote this approach by ``share-enc''. Finally, (iii) each emotion has a dedicated encoder and decoder. We denote this approach by ``share-none''.  We evaluated all three approaches and summarized the results in Table~\ref{tab:arch}.

We report the BLEU score~\cite{papineni2002bleu} and \ac{UER}. \ac{UER} measures relative edit-distance between the target unit sequence and the generated unit sequence. We additionally report the \ac{WER} and \ac{PER} metrics extracted using a pre-trained \ac{SOTA} \ac{ASR} system~\footnote{We use a \textsc{base} wav2vec 2.0 phoneme detection model trained on LibriSpeech-960h with CTC loss from scratch.}. Results are reported on emotions converted to neutral: \{amused, angry, sleepy, disgusted\}-neutral. The BLEU and \ac{UER} metrics are computed using all emotion pairs from Neutral and to Neutral. 

\begin{table}[h!]
    \centering
    \begin{tabular}{c|c|c|c|c}
        \toprule
        Architecture & BLEU & \ac{UER} & \ac{WER} & \ac{PER} \\
        \toprule
         Share-all  & 31.97 & 42.56 & 25.47 & 26.44 \\ \hline
         Share-enc  & 32.22 & \bf 41.16 & \bf 23.91 & \bf 25.41 \\ \hline
         Share-none & \bf 32.37 & 41.38 & \bf 23.91 & 25.95 \\
         \bottomrule
    \end{tabular}
    \caption{Evaluation of three weight-sharing schemes for the translation model.}
    \vskip -1em
    \label{tab:arch}
\end{table}

It can be seen that the share-enc and share-none architectures are comparable, while the share-all configuration is inferior.
Although both share-enc and share-none configurations are similar in terms of lexical reconstruction, in our listening tests share-none generated more expressive speech.

\paragraph{F0 estimation module}
We evaluate the F0 estimation model using the \ac{MAE} between ground-truth F0 and predicted F0. In this experiment, we explore a number of configurations for training such an estimator. Specifically, we evaluate different binning strategies, normalization methods and prediction rules. For binning strategies, we explore adaptive binning vs. uniform binning. Under normalization, we explore no normalization, mean normalization and mean \& standard deviation normalization\footnote{The mean and standard deviation are computed using the F0 values per speaker.}. Finally, in addition to the weighted-average prediction rule described in Section~\ref{sec:speech_modeling}, we also evaluate an argmax prediction rule where the highest scoring bin is predicted. Results are summarized in Table~\ref{tab:f0_log}, ``Log'' denotes applying the logarithm function before normalization. 

Results suggest that the weighted-average prediction rule is preferable to argmax, especially when used in conjunction with adaptive binning. This can be explained by large-range bins in the adaptive case, leading to larger \ac{MAE} when selecting a single bin using the argmax operator. Although adaptive quantization reaches the best performance, under specific settings uniform quantization can reach comparable results. For normalization, it is preferable to normalize, the specific normalization method has little impact on performance. 

\begin{table}[t!]
    \centering
    \resizebox{0.9\columnwidth}{!}{%
    \begin{tabular}{c|c|c|c|c}
        \toprule
        Log & Quantization & Norm.  & Prediction & MAE \\
        \toprule
         \ding{55} &  uniform     & \ding{55} & argmax & 83.51 \\ \hline
         \ding{55} &  uniform     & \ding{55} & w-avg  & 44.92 \\ \hline
         \ding{55} &  uniform     & mean      & argmax & 53.74 \\ \hline
         \ding{55} &  uniform     & mean      & w-avg  & 35.63 \\ \hline
         \ding{55} &  uniform     & mean-std  & argmax & 63.01 \\ \hline
         \ding{55} &  uniform     & mean-std  & w-avg  & 35.69 \\ \hline
         \ding{55} &  adaptive    & \ding{55} & argmax & 129.7 \\ \hline
         \ding{55} &  adaptive    & \ding{55} & w-avg  & 45.21 \\ \hline
         \ding{55} &  adaptive    & mean      & argmax & 127.8 \\ \hline
         \ding{55} &  adaptive    & mean      & w-avg  & 36.06 \\ \hline
         \ding{55} &  adaptive    & mean-std  & argmax & 155.7 \\ \hline
         \ding{55} &  adaptive    & mean-std  & w-avg  & \bf 35.38 \\ \hline
         
         \ding{51} &  uniform     & \ding{55} & argmax & 62.69 \\ \hline
         \ding{51} &  uniform     & \ding{55} & w-avg  & 67.40 \\ \hline
         \ding{51} &  uniform     & mean      & argmax & 52.24 \\ \hline
         \ding{51} &  uniform     & mean      & w-avg  & 63.67 \\ \hline
         \ding{51} &  uniform     & mean-std  & argmax & 50.95 \\ \hline
         \ding{51} &  uniform     & mean-std  & w-avg  & 51.25 \\ \hline
         \ding{51} &  adaptive    & \ding{55} & argmax & 127.9 \\ \hline
         \ding{51} &  adaptive    & \ding{55} & w-avg  & 67.40 \\ \hline
         \ding{51} &  adaptive    & mean      & argmax & 110.4 \\ \hline
         \ding{51} &  adaptive    & mean      & w-avg  & 54.21 \\ \hline
         \ding{51} &  adaptive    & mean-std  & argmax & 144.9 \\ \hline
         \ding{51} &  adaptive    & mean-std  & w-avg  & 51.25 \\ \hline
         \bottomrule
    \end{tabular}
    }
    \caption{Evaluation of different F0 estimation configurations. The MAE is reported for voiced frames only.}
    \vskip -1em
    \label{tab:f0_log}
\end{table}

\paragraph{Duration prediction module.}
We evaluate the duration prediction models using the \ac{MAE} between target and predicted durations. For a more complete analysis, we also report the accuracy using thresholds of 0ms, 20ms and 40ms. We explore a \ac{CNN} duration predictor similarly to \cite{fastspeech2} and three n-gram based models. The results are summarized in Table~\ref{tab:dur}. As expected, the \ac{CNN} outperforms n-gram models, with $\sim$94\% accuracy when considering a tolerance level of 40ms.
\begin{table}[t!]
    \centering
    \resizebox{0.99\columnwidth}{!}{%
    \begin{tabular}{c|c|c|c|c}
        \toprule
        Model & MAE & Acc@0ms  & Acc@20ms & Acc@40ms \\
        \toprule
         \ac{CNN}  & \bf 0.77 & \bf 51.12 & \bf 86.24 & \bf 94.08 \\ \hline
         1-gram    & 1.47 & 29.41 & 67.34 & 83.26 \\ \hline
         3-gram    & 1.16 & 36.78 & 76.04 & 88.32 \\ \hline
         5-gram    & 1.02 & 37.32 & 81.36 & 90.46 \\
         \bottomrule
    \end{tabular}}
    \caption{Evaluation of four duration prediction models. CNN architecture is as described in~\cite{fastspeech2}.}
    \vspace{-0.2cm}
    \label{tab:dur}
\end{table}

\begin{figure}[h!]
\centering
  \centering
  \includegraphics[width=0.8\linewidth]{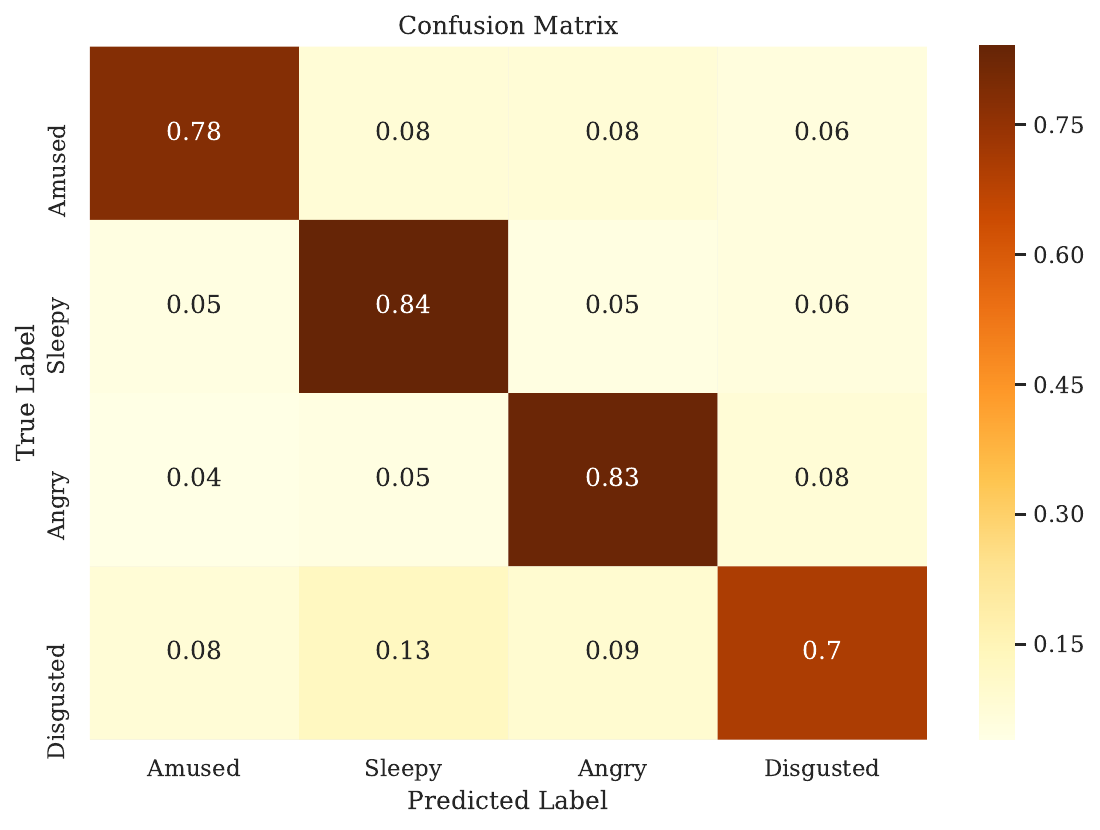}  
  \caption{Ground Truth Recordings}
  \label{fig:gt_conf}
\end{figure}

\begin{figure}[h!]
  \centering
  \includegraphics[width=0.8\linewidth]{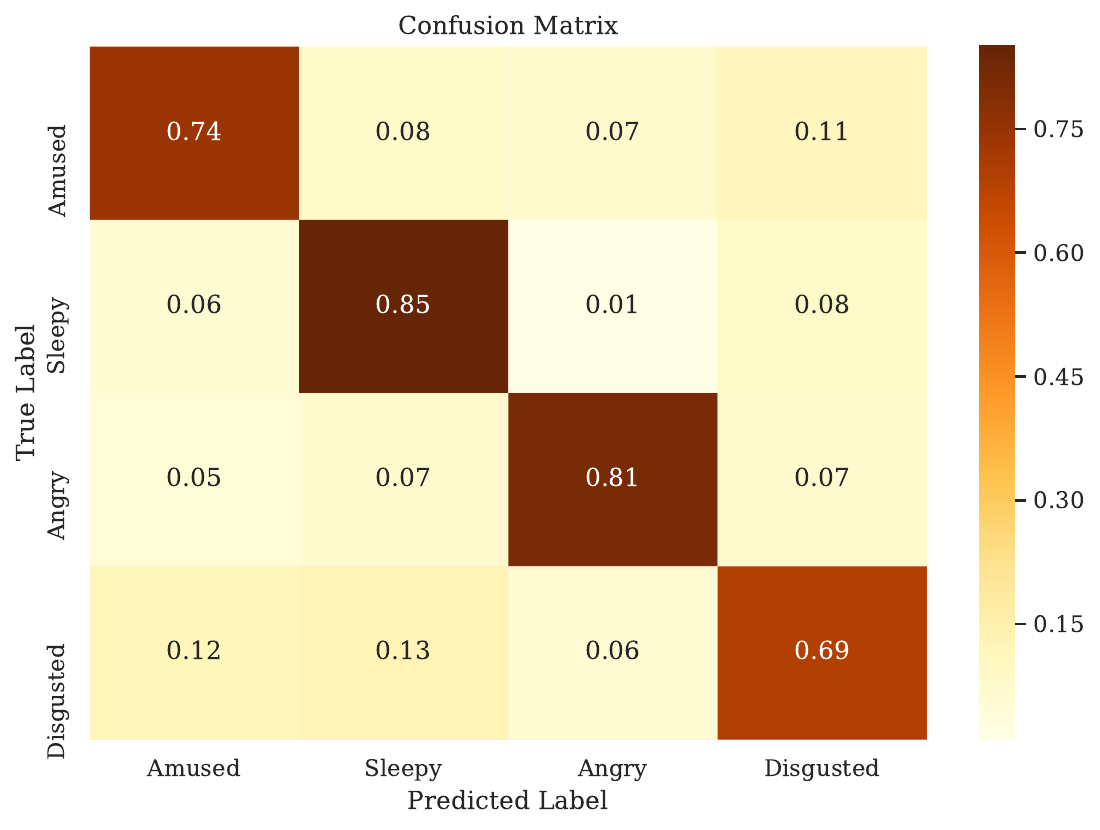}  
  \caption{$\oursp$}
  \label{fig:ours_conf}
\end{figure}

\begin{figure}[h!]
  \centering
  \includegraphics[width=0.8\linewidth]{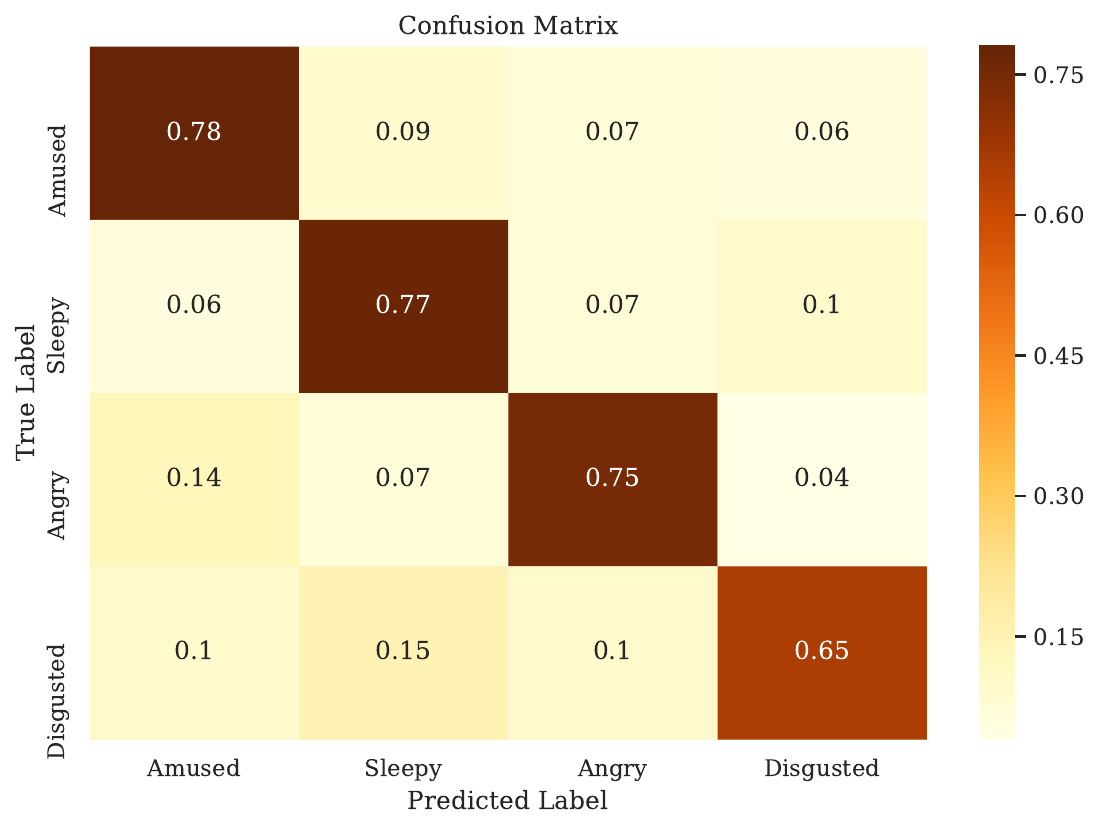}  
  \caption{$\oursn$}
  \label{fig:ours_m_conf}
\end{figure}

\begin{figure}[h!]
  \centering
  \includegraphics[width=0.8\linewidth]{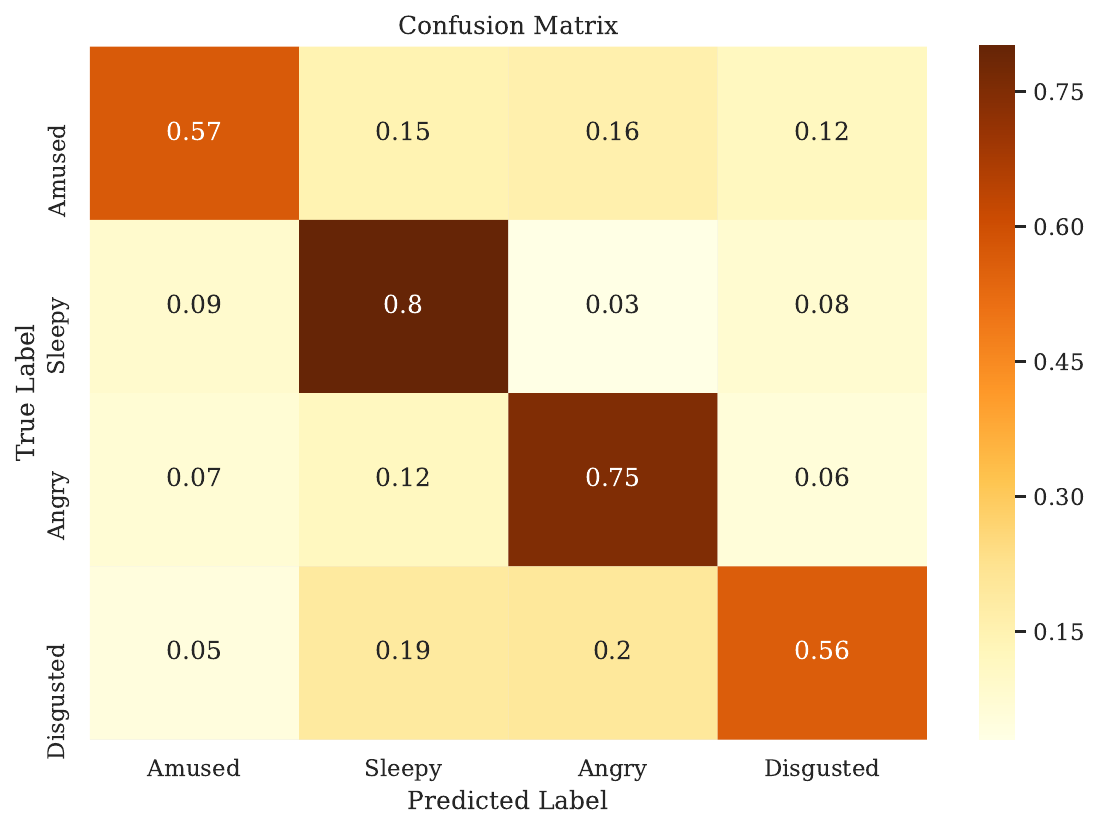}  
  \caption{Seq2Seq-EVC}
  \label{fig:s2s_conf}
\end{figure}

\begin{figure}[h!]
  \centering
  \includegraphics[width=0.8\linewidth]{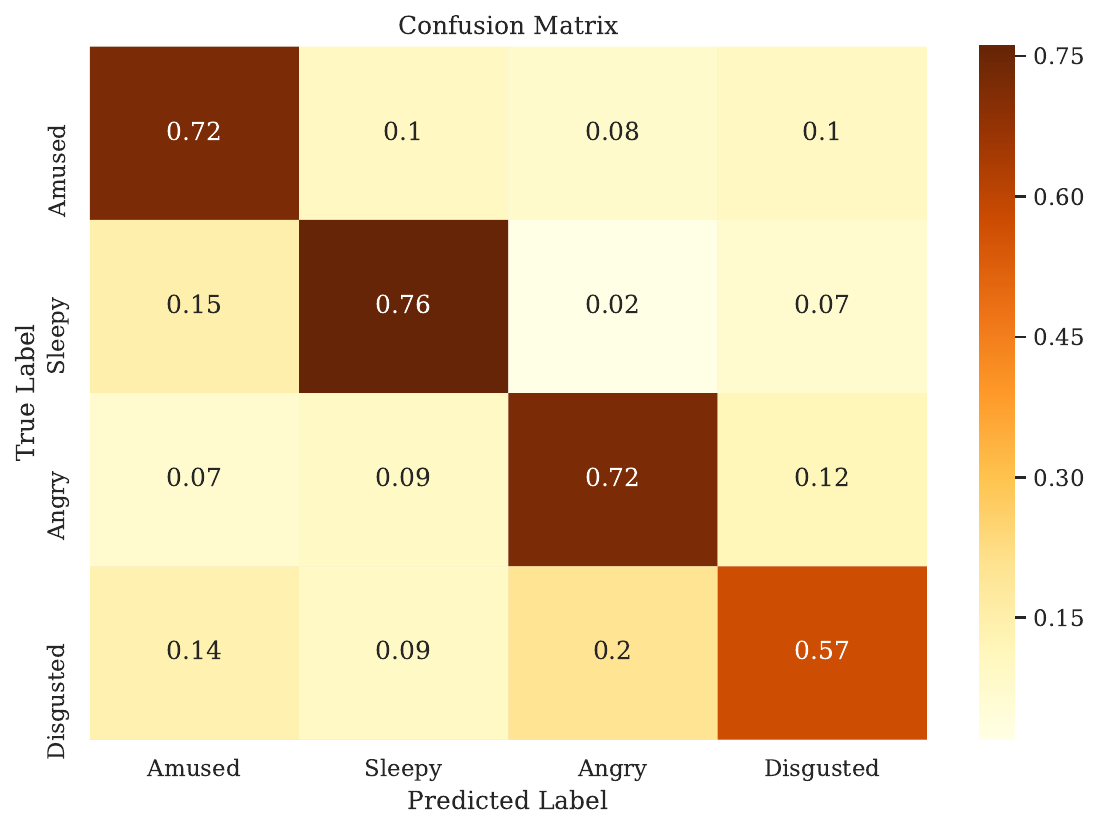}  
  \caption{Tacotron2}
  \label{fig:tac_conf}
\end{figure}

\begin{figure}[h!]
  \centering
  \includegraphics[width=0.8\linewidth]{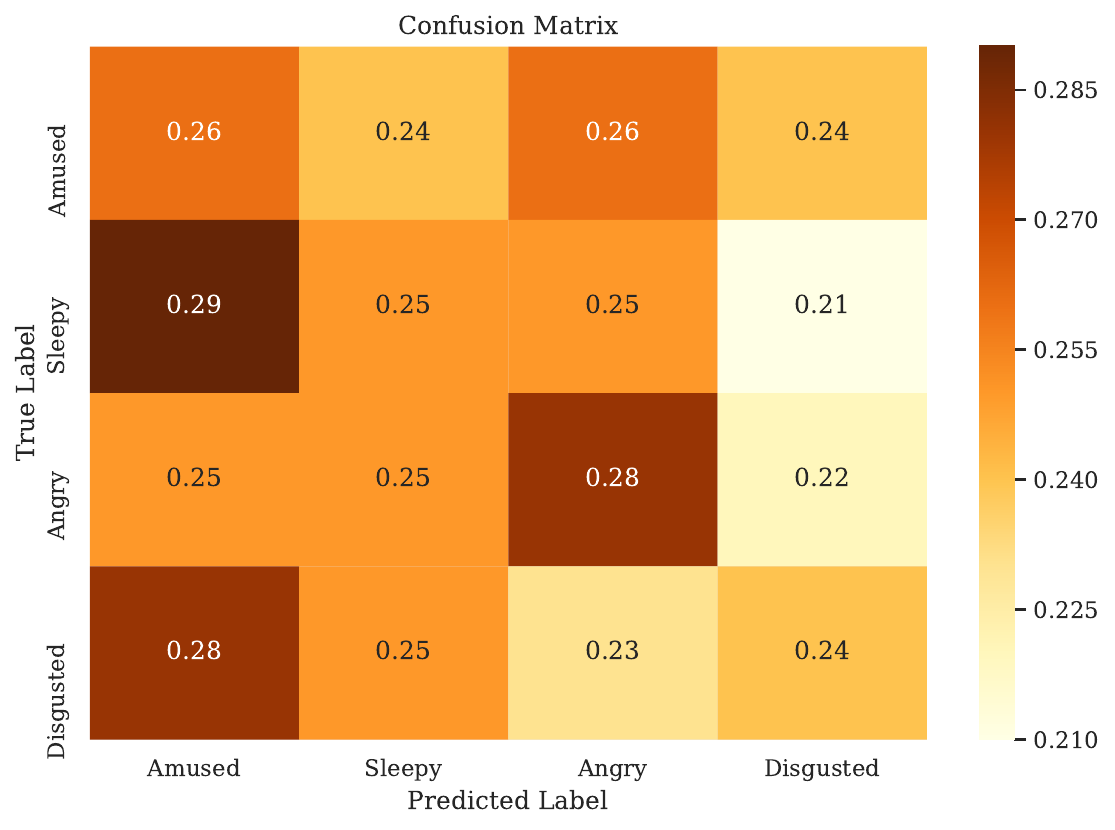}  
  \caption{VAW-GAN}
  \label{fig:vaw_conf}
\end{figure}

\paragraph{\ac{eMOC} Confusion Matrices}
We provide the \ac{eMOC} confusion matrices for samples generated by each of the evaluated models and the ground truth recordings. See Figures~\ref{fig:gt_conf},~\ref{fig:ours_conf},~\ref{fig:ours_m_conf},~\ref{fig:s2s_conf},~\ref{fig:tac_conf},~\ref{fig:vaw_conf}. We additionally provide the full eMOC results for the many-to-many setting in the Table~\ref{sup:emov_all2all_full}.

\subsection{Subjective Evaluation} For the crowed-sourced subjective evaluations we computed the MOS and eMOC metrics. For the MOS metric raters were asked: ``Rate the quality and naturalness of the given speech utterance on a scale of 1 to 5 (1 being of low quality and naturalness and 5 being of high quality and naturalness)''. For the eMOC raters were asked: ``Select the emotion for the given emotion categories that best suits the given speech utterance''.

All raters are native English speakers located in the United-States.

\begin{table}[ht!]
\centering
\resizebox{0.95\columnwidth}{!}{%
\begin{tabular}{l|l|c|c|c|c|c}
\toprule
Source & Target & Amused & Angry & Disgusted & Sleepy & Neutral \\
\toprule
\multirow{5}{*}{Amused}    & Amused    & -      & -              & -                  & -               & -                \\
                           & Angry     & 0.06            & \textbf{0.81}  & 0.04               & 0.04            & 0.05             \\
                           & Disgusted & 0.04            & 0.04           & \textbf{0.77}      & 0.01            & 0.14             \\
                           & Sleepy    & 0.06            & 0.02           & 0.03               & \textbf{0.85}   & 0.04             \\
                           & Neutral   & 0.06            & 0.07           & 0.09               & 0.03            & \textbf{0.75}    \\
\midrule
\multirow{5}{*}{Angry}     & Amused    & \textbf{0.83}   & 0.07           & 0.04               & 0.02            & 0.04             \\
                           & Angry     & -               & -              & -                  & -               & -                \\
                           & Disgusted & 0.05            & 0.12           & \textbf{0.59}      & 0.07            & 0.17             \\
                           & Sleepy    & 0.04            & 0.06           & 0.03               & \textbf{0.82}   & 0.05             \\
                           & Neutral   & 0.06            & 0.08           & 0.04               & 0.03            & \textbf{0.79}    \\
\midrule
\multirow{5}{*}{Disgusted} & Amused    & \textbf{0.71}   & 0.08           & 0.1                & 0.05            & 0.06             \\
                           & Angry     & 0.03            & \textbf{0.78}  & 0.05               & 0.09            & 0.05             \\
                           & Disgusted & -               & -              & -                  & -               & -                \\
                           & Sleepy    & 0.09            & 0.06           & 0.05               & \textbf{0.72}   & 0.08             \\
                           & Neutral   & 0.08            & 0.02           & 0.1                & 0.03            & \textbf{0.77}    \\
\midrule
\multirow{5}{*}{Sleepy}    & Amused    & \textbf{0.63}   & 0.02           & 0.12               & 0.14            & 0.09             \\
                           & Angry     & 0.06            & \textbf{0.77}  & 0.03               & 0.09            & 0.05             \\
                           & Disgusted & 0.03            & 0.07           & \textbf{0.66}      & 0.13            & 0.11             \\
                           & Sleepy    & -               & 0              & -                  & -               & -                \\
                           & Neutral   & 0.05            & 0.08           & 0.12               & 0.12            & \textbf{0.63}    \\
\midrule
\multirow{5}{*}{Neutral}   & Amused    & \textbf{0.7}    & 0.08           & 0.06               & 0.08            & 0.08             \\
                           & Angry     & 0.06            & \textbf{0.77}  & 0.06               & 0.07            & 0.04             \\
                           & Disgusted & 0.08            & 0.07           & \textbf{0.63}      & 0.1             & 0.12             \\
                           & Sleepy    & 0.06            & 0.02           & 0.03               & \textbf{0.79}   & 0.1              \\
                           & Neutral   & -               & -              & -                  & -               & -             \\
\bottomrule
\end{tabular}}
\caption{Subjective evaluation for any-to-any emotion conversion. The ``source'' and ``target'' columns represent the source and target emotion respectively. Each emotion column (Amused, Angry, Disgusted, Sleepy and Neutral columns) represent the percentage of human raters that chose the said emotion when converting from source to target emotion.}
\label{sup:emov_all2all_full}
\end{table}

\section{Risks \& Limitations}
The main risks in the proposed approach (as in any generative model) is the development of a high quality and natural speech synthesis model. Such technology might be used to alter emotion in speech recordings. To deal with that we limit the number of speakers that can be synthesized with the proposed approach using a look-up-table. Another potential risk of the proposed method which is also one of its limitations is that the generated speech content is not always perfect, hence might lead to wrong pronunciations. As described in the results section, the WER of the proposed system is $\sim$27.92 which means we still have a big room for improvement. Another limitation of the current approach is the need for parallel corpus to convert between different emotions.

\end{document}